\documentclass[10pt,twocolumn,letterpaper]{article}

\usepackage[review]{cvpr}      

\usepackage[dvipsnames]{xcolor}

\definecolor{cvprblue}{rgb}{0.21,0.49,0.74}
\usepackage[pagebackref,breaklinks,colorlinks,citecolor=cvprblue]{hyperref}
\usepackage{graphicx}

\title{Watermarking in Diffusion Model: Gaussian Shading with Exact Diffusion Inversion via Coupled Transformations (EDICT)
 }

\author{Krishna Panthi\\
School of Computing,
Clemson University\\
Clemson, South Carolina, USA \\
{\tt\small kpanthi@clemson.edu}}

\begin{document}
\maketitle
\begin{abstract}
This paper introduces a novel approach to enhance the performance of Gaussian Shading, a prevalent watermarking technique, by integrating the Exact Diffusion Inversion via Coupled Transformations (EDICT) framework. While Gaussian Shading traditionally embeds watermarks in a noise latent space, followed by iterative denoising for image generation and noise addition for watermark recovery, its inversion process is not exact, leading to potential watermark distortion. We propose to leverage EDICT's ability to derive exact inverse mappings to refine this process. Our method involves duplicating the watermark-infused noisy latent and employing a reciprocal, alternating denoising and noising scheme between the two latents, facilitated by EDICT. This allows for a more precise reconstruction of both the image and the embedded watermark. Empirical evaluation on standard datasets demonstrates that our integrated approach yields a slight, yet statistically significant improvement in watermark recovery fidelity. These results highlight the potential of EDICT to enhance existing diffusion-based watermarking techniques by providing a more accurate and robust inversion mechanism. To the best of our knowledge, this is the first work to explore the synergy between EDICT and Gaussian Shading for digital watermarking, opening new avenues for research in robust and high-fidelity watermark embedding and extraction.
\end{abstract}    
\section{Introduction}
\label{sec:intro}

AI-generated images have become increasingly prevalent. Just a few years ago, it was very expensive to generate realistic-looking images using computers. But now, in 2024, they have become the norm. It’s becoming harder and harder to tell whether an image is real or AI-generated. Images have the capability to deceive people who view them. When a person sees an image, it leaves a long-lasting impression on their mind. Unlike text, understanding an image does not require much time. Consequently, it has become far easier to spread misinformation.

For example, an AI-generated image was used to report fake news about an explosion outside the Pentagon on the social media platform X. This caused stock prices to drop temporarily. This highlights the importance of identifying whether a given image was generated by AI or not. Furthermore, being able to identify who generated the image could help deter people from creating such misleading images.

Diffusion models are the current state-of-the-art approach to generating images. Most modern advanced systems, such as MidJourney, DALL-E, and Stable Diffusion, all use some form of diffusion models to generate images. Diffusion models generate images by iteratively denoising input Gaussian noise in the direction of the prompt. If we can embed some form of secret watermark into the image generation process, we could later recover it to determine whether the image was generated by a specific model or even trace its creator. Gaussian Shading \cite{yang2024gaussian} is one such technique that embeds a watermark into the noise itself, which is used to generate the image at runtime. The same model can then be used to recover the noise and, ultimately, the watermark for detection and traceability. This method works with diffusion models that use Denoising Diffusion Implicit Models (DDIMs)\cite{song2020denoising}, which employ non-Markovian methods of image generation. The challenge, however, is that the initial noise must be recoverable from the image for this method to be successful.

Even though DDIMs are non-Markovian models, it is still not possible to fully recover the initial noise. DDIM inversion for images is unstable because it relies on local linearization assumptions. These assumptions result in the propagation of errors, leading to incorrect image reconstruction and loss of content. This also affects the embedded watermark, making it harder to recover. This is especially true when images are manipulated after being generated.

Exact Diffusion Inversion via Coupled Transformations (EDICT)\cite{wallace2022edict} is an inversion method inspired by affine coupling layers. EDICT enables mathematically exact inversion of real and model-generated images by maintaining two coupled noise vectors that invert each other in an alternating fashion. To the best of our knowledge, this work presents the first implementation of the EDICT framework in conjunction with Gaussian Shading for digital watermarking. Our experimental results indicate that this novel integration leads to a modest improvement in watermark recovery performance, suggesting the viability of EDICT for enhancing existing watermarking techniques.

Both EDICT and Gaussian Shading do not require model training or fine-tuning. We observed a slight increase in the performance of Gaussian Shading when EDICT is implemented.

Project Website: \hyperlink{https://krishnapanthi.com/projects/gaussian-shading-with-edict/}{krishnapanthi.com}

\section{Related Work}
\label{sec:intro}
This work builds on two key contributions. The first is Gaussian Shading \cite{yang2024gaussian}, a watermarking technique that embeds a unique watermark directly into latent noise. It works with DDIM based latent diffusion models. A model owner can integrate such a watermark, the resulting images carry a signature that proves it's authenticity as well as it can be traced back to user who generated it. If malicious users generate harmful or misleading content, the traceable watermark can identify the user, helping to hold the user accountable and discouraging misuse. 

Let us consider c x h x w is the dimension of the latent. In such a scenario, the watermark capacity becomes l x c x h x w. To enhance robustness, the watermark is represented by $\frac{1}{f_{hw}}$ of the height and width, $\frac{1}{f_{c}}$ of the channel. The watermark is then expanded to fill the latent by $f_{hw}^2 \cdot f_c$. The watermark has capacity $l \times \frac{c}{f_c} \times \frac{h}{f_{hw}} \times \frac{w}{f_{hw}} $. Let $s^d$ be the diffused watermark. The diffused watermark is encrypted with a key K. The resulting message m is a randomized binary string and has a uniform distribution, which is a property of a good encryption algorithm. $l$ refers to the number of bits represented by each dimension which results in an integer $y$ in the range [0, 1,...,$2^{l-1}$]. 

To embed y, the standard normal distribution $N(0,1)$ is divided into $2^l$ equal cumulative probability portions. For implementation with l=1, this means for each bit value of 0, we sample from a truncated standard normal distribution $N(0,1)$ in the range $[-\infty, 0)$, and for each bit value of 1, we sample from a truncated standard normal distribution $N(0,1)$ in the range $[0, \infty)$.

The second work is Exact Diffusion Inversion via Coupled Transformations (EDICT) \cite{wallace2022edict}. EDICT is inspired by affine coupling layers (ACL) from invertible neural networks \cite{dinh2015nicenonlinearindependentcomponents, dinh2017densityestimationusingreal}. This process ensures a mathematically exact inversion of diffusion processes by utilizing two coupled latents that invert each other in an alternating manner.  Unlike standard DDIM inversion, which is prone to errors and leads to imperfect reconstruction, EDICT guarantees the recovery of the original noise latent without approximation.

The way it works is in a normal diffusion one latent is used for denoising and noising in the forward and reverse process, however in EDICT in each time step, two latents are used, also called coupled latents where one latent is used to noise or denoise other in an alternative way. After that they are mixed together. And the same process is repeater for all time steps. Given $x_t$ and $y_t$ are the latents at time t, and $0 \le p \le 1$ is mixing factor, we have the next latents $x_{t-1}$, $y_{t-1}$ calculated in denoising process as

\begin{align*}
x^{inter}_t &= \text{denoise}(x_t, t, y_t) \\
y^{inter}_t &= \text{denoise}(y_t, t, x^{inter}_t) \\
x_{t-1} &= p \cdot x^{inter}_t + (1-p)\cdot y^{inter}_t \\
y_{t-1} &= p \cdot y^{inter}_t + (1-p)\cdot x_{t-1}
\end{align*}

and the deterministic noising inversion process is

\begin{align*}
y^{inter}_{t+1} &= (y_t - (1-p)\cdot x_t)/p \\
x^{inter}_{t+1} &= (x_t - (1-p)\cdot y^{inter}_{t+1})/p \\
y_{t+1} &= \text{addnoise}(y^{inter}_{t+1}, t+1, x^{inter}_{t+1}) \\
x_{t+1} &= \text{addnoise}(x^{inter}_{t+1}, t+1, y_{t+1})
\end{align*}

$denoise(x, t, y)$, given a latent variable $x$ at time step $t$ and another latent $y$, it applies the model’s reverse diffusion step. It typically employs a neural network conditioned on $y$ and time $t$ to estimate the denoised version of $x$, reducing the noise from the current latent representation.

$addnoise(x, t, y)$, given a latent variable $x$ at time step $t$ and another latent $y$, it applies the forward diffusion step. It adds a controlled amount of noise to $x$, according to a predefined noise schedule, optionally conditioned on $y$ and the current time step $t$. This operation effectively inverts the denoising process, enabling transitions between different time steps.

In practice, the order in which the x and y series are calculated are alternated at each time step in order to symmetrize the process with respect to both sequences.

Combining these approaches is novel because it enhances the watermarking method’s reliability under a wide range of transformations and image manipulations. While Gaussian Shading can embed a watermark, and DDIM-based methods can partially invert the process, EDICT’s exact inversion makes it possible to precisely extract the embedded watermark from generated images. Our extension focuses on leveraging EDICT to maintain higher fidelity in watermark recovery, something previous work did not fully address.

\section{Approach}
\label{approach}

This paper introduces a novel approach that extends the Gaussian Shading technique by integrating the EDICT framework to facilitate the derivation of exact inverse mappings. Gaussian Shading traditionally involves embedding a watermark within a singular noise latent. This watermark-infused latent is then subjected to an iterative denoising process, ultimately converging upon a clean, denoised latent representation. Concurrently, an image latent undergoes an iterative process of noise addition, culminating in a noise-infused latent from which the embedded watermark can be extracted.

The proposed enhancement leverages EDICT's capabilities by initially duplicating the watermark-infused noisy latent, thereby generating two distinct latent representations. These latents are then engaged in a reciprocal, alternating denoising process. Specifically, each latent iteratively refines the other, progressively removing noise in a coordinated manner. Ultimately, one of these refined latents is utilized for image generation, while the other is discarded.

In the inverse process, a denoised latent representation of the input image is initially obtained via the encoder network. This latent is then duplicated, creating two identical latents. These latents are subsequently employed in a reciprocal noise generation process, effectively reversing the steps of the denoising phase. In this alternate flow, each latent contributes to the noise profile of its counterpart. Finally, one of the resulting noisy latents is used to recover the embedded watermark. This bidirectional procedure, facilitated by EDICT, ensures a more precise inversion process, leading to a significantly less distorted watermark extraction.

The algorithmic details of this combined approach, encompassing both the forward (watermark embedding and image generation) and inverse (watermark extraction) processes, are elucidated in the pseudo-code presented below.
\\
\textbf{Input:}\\
$W$ : Watermark to be Embedded\\
$Decoder(\cdot)$: Image generation model\\
$Encoder(\cdot)$: Latent space encoding model\\
$EDICT\_denoise(\cdot,\cdot)$: EDICT based bidirectional denoising process.\\
$EDICT\_reverse(\cdot, \cdot)$: Reverse EDICT process.\\
$T$: Total timesteps in the denoising process.\\
\\
\textbf{Output:}\\
Recovered watermark: $\hat{W}$.\\
Generated Image: $I$.\\
\\
\textbf{Algorithm:}\\
\\
\textbf{
Forward Process (Denoising)}
\begin{enumerate}
\item $x_t \xleftarrow{} EmbedWatermark(W)$
\item $y_t \xleftarrow{} Duplicate(x_1)$
\item $(x_0, y_0) \xleftarrow{} EDICT\_denoise(x_t,t,y_t)$
\item $I \xleftarrow{} Decoder(x_0)$\\
\end{enumerate}
 \textbf{Reverse Process (Adding Noise)}

\begin{enumerate}
\item $x_0 \xleftarrow{} Encoder(I)$
\item $y_0 \xleftarrow{} Duplicate(x_0)$
\item $(x_t, y_t) \xleftarrow{} EDICT\_reverse(x_0, t, y_0)$
\item $\hat{W} \xleftarrow{} RecoverWatermark(x_t) $
\item $Compare(W, \hat{W})$\\
\end{enumerate}

In essence, we use two coupled latents (x and y) that guide each other’s denoising in the forward pass, and then use the reverse operation to reintroduce noise step-by-step.

\section{Experiments and Results}
\label{experiments}

\begin{figure*}
  \centering
  \begin{tabular}{ccccc}
    \includegraphics[width=0.16\textwidth]{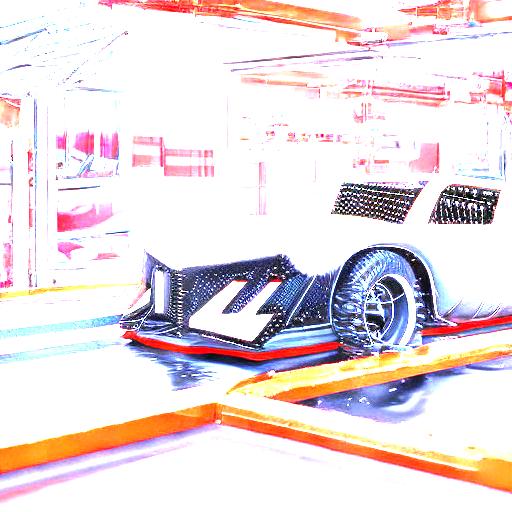} &
    \includegraphics[width=0.16\textwidth]{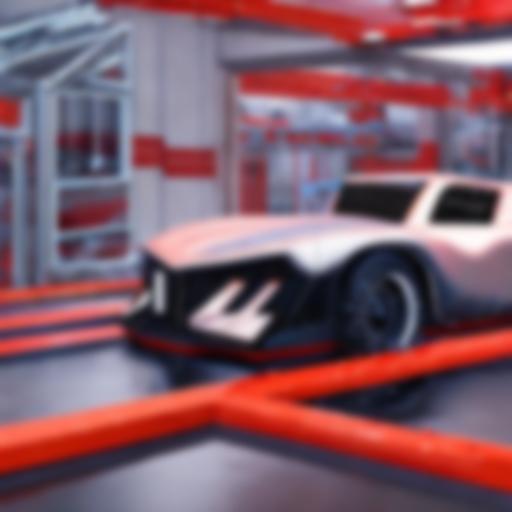} &
    \includegraphics[width=0.16\textwidth]{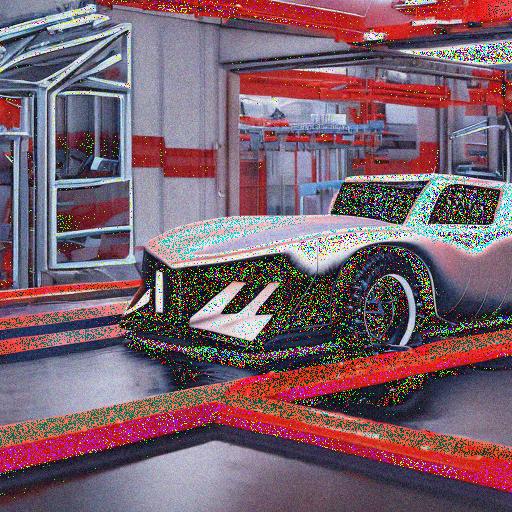} &
    \includegraphics[width=0.16\textwidth]{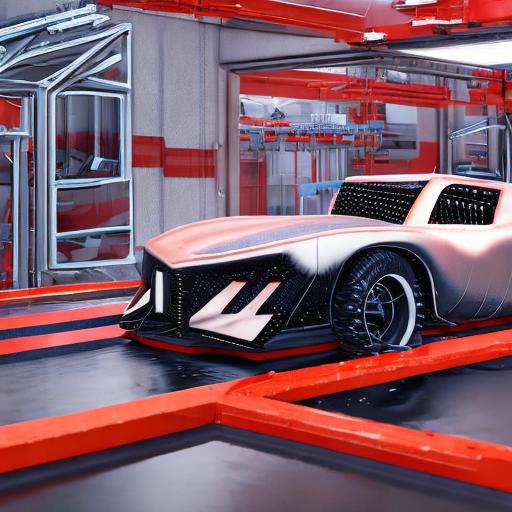} &
    \includegraphics[width=0.16\textwidth]{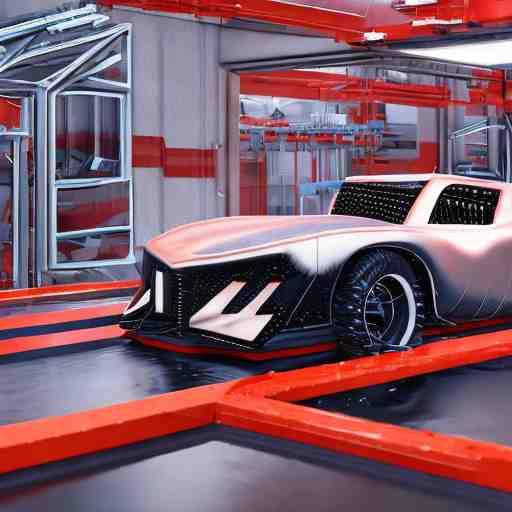} \\
     (a) & (b) & (c) & (d) & (e)\\
    \includegraphics[width=0.16\textwidth]{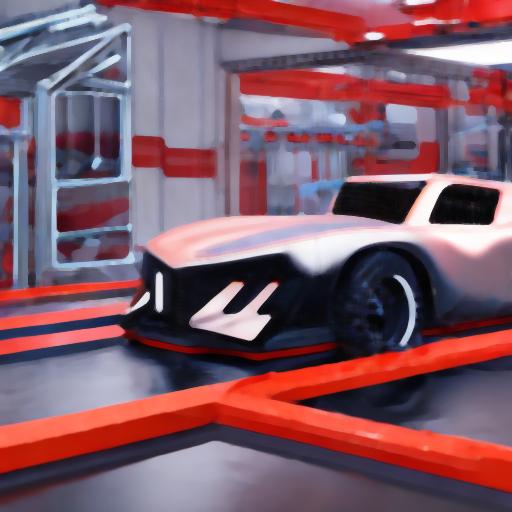} &
    \includegraphics[width=0.16\textwidth]{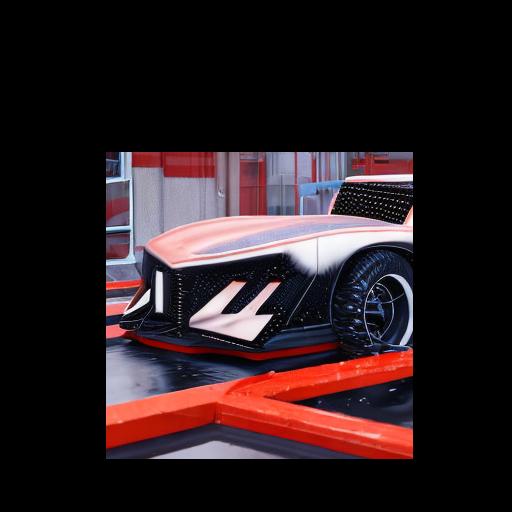} &
    \includegraphics[width=0.16\textwidth]{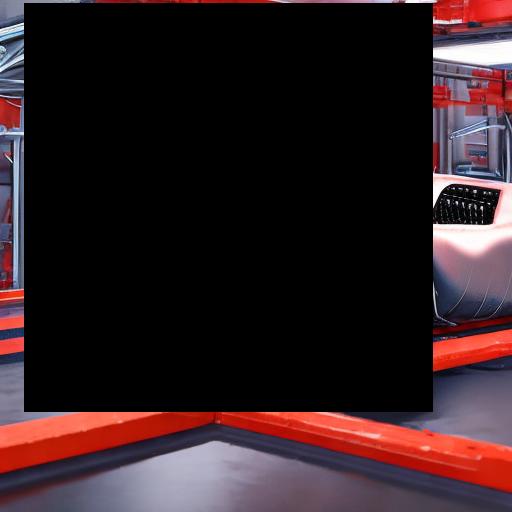} &
    \includegraphics[width=0.16\textwidth]{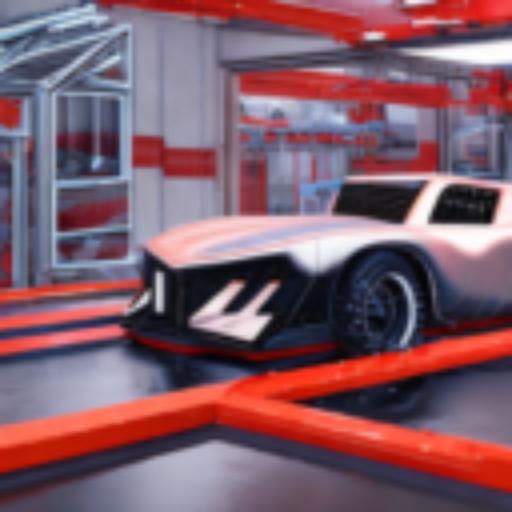} &
    \includegraphics[width=0.16\textwidth]{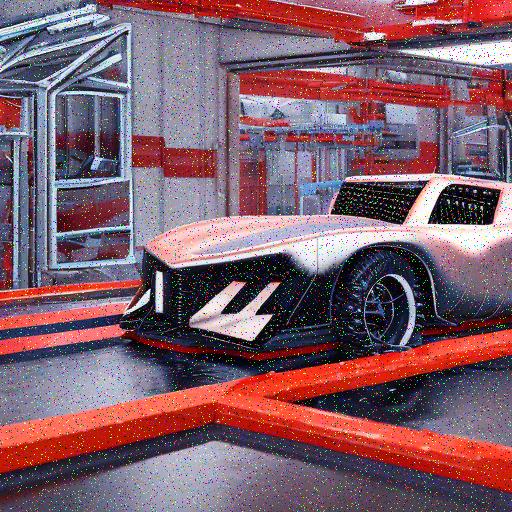} \\
    (f) & (g) & (h) & (i) & (j)\\
  \end{tabular}
  \caption{Example of images manipulated. (a) Brightness, factor = 6 (Color Jitter), (b) Gaussian Blur, r=4 (GauBlur),
    (c) Gaussian Noise, $\mu = 0, \sigma = 0.05$ (GauNoise), (d) Identity,
    (e) JPEG, QF=25, (f) Median Filter, k=7 (MedBlur), (g) 60\% area random crop,
    (h) 80\% area random drop,  (i) 25\% Resize and restore (Resize), (j) Salt and Pepper Noise, p = 0.05 (S\&PNoise) }
  \label{fig:multiple_images}
\end{figure*}

\begin{figure*}
  \centering
  \hfill
  \includegraphics[scale=0.62]{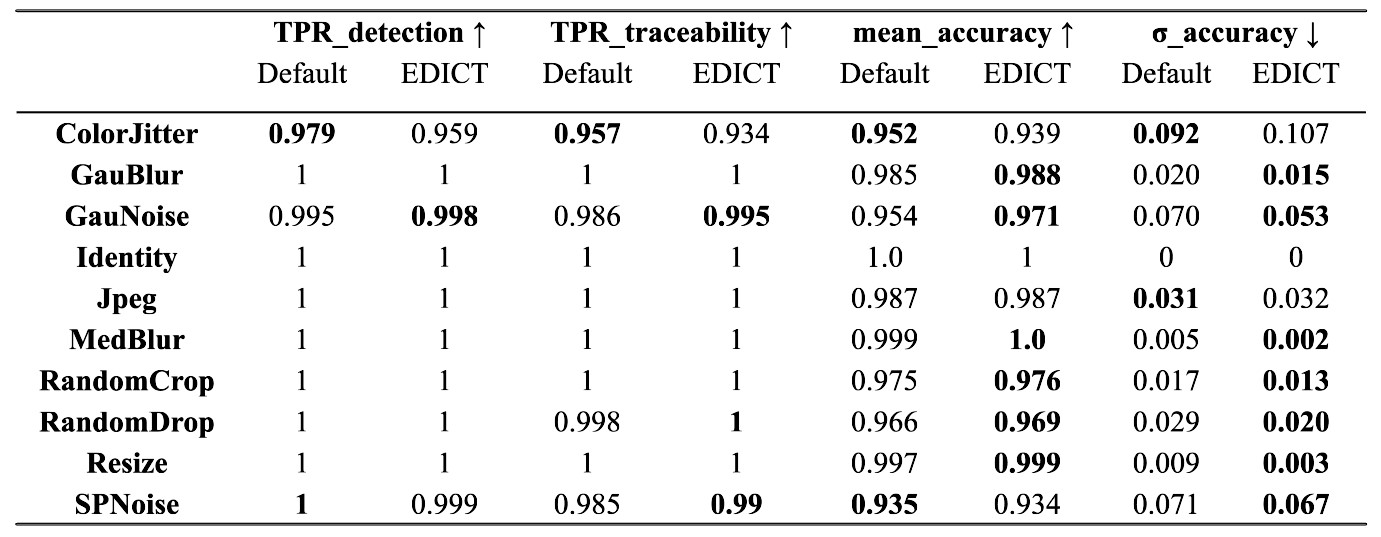}
  \caption{The table shows the results obtained by testing our method against the baseline. It demonstrates that when EDICT is used, performance improves or remains consistent across all image manipulation methods, except when brightness is increased (ColorJitter) and when Salt and Pepper noise is added.}
  \label{fig:short}
\end{figure*}

We evaluated our approach by comparing Gaussian Shading’s watermark recovery performance with and without EDICT. The experiment was conducted with Stable Diffusion 2.1 \cite{Rombach_2022_CVPR} provided by huggingface. The size of the generated images is $512 \times 512$, and the latent space dimension is $4 \times 64 \times 64$. During inference we emply the prompt from Stable-Diffusion-Prompt with a guidance scale of 7.5 similar to the Gaussian Shading paper. We sample 50 steps using DDIMSolver \cite{song2022denoisingdiffusionimplicitmodels}. 50 steps of DDIM inversion was performed. For the experiment the settings of Gaussian Shading were $f_c = 1$, $f_{hw} = 8$, $l = 1$, resulting in an actual capacity of 256 bits.  For robustness, we subjected generated images to nine different noise as shown in Figure 1. 

All experiments are conducted using the PyTorch 2.5.1 framework, running on a single A100 GPU.

\textbf{Evaluation metrics:} For detection, we calculate the true positive rate (TPR) corresponding to a fixed false positive rate (FPR). For traceability, bit accuracy is calculated.

\subsection{Performance}

For detection, we consider Gaussian Shading a single bit watermark, with a fixed watermark. We approximate a fixed FPR of $10^{-6}$, calculate the corresponding threshold $\tau$ which comes out to be approximately equal to 78\% of bits being correct and test the TPR on 1000 images. As seen in the results table, when using EDICT the TPR reduces for ColorJitter and SPNoise, increases for GauNoise and remains same for all others. 

For traceability, Gaussian Shading serves as a multi-bit watermark. In the experiment, we consider that 1,000 users generate 10 images each with a watermark, resulting to 10,000 watermarked images. During testing we calculate the threshold $\tau$ to control the FPR at $10^{-6}$ which approximately equals 88\% of bits being correct. As seen in the results table, when using EDICT, the traceability improves for GauNoise, Random Drop and SPNoise but reduces for ColorJitter, remains same for all others. 

We also calculated the bit accuracy for both EDICT implementation and baseline to compare the results. As seen in the table, EDICT improves bit accuracy for 7 out of 9 noise addition methods. The accuracy goes slightly down for Brightness and S \& P Noise.
\section{Conclusions}

We integrated EDICT’s exact inversion capability with Gaussian Shading’s watermarking method to improve the fidelity of watermark recovery in diffusion-generated images. Our experiments show that while the gains are not always dramatic, EDICT does provide a more stable inversion process, reducing errors in watermark recovery, particularly in challenging transformations. The method does not require model retraining or finetuning, which makes it easily adaptable.

Limitations include the complexity of implementing exact inversion and the computational overhead introduced by EDICT’s coupled transforms. This method is 2 x slower than the baseline. In the future we will try to improve the performance of EDICT. We will also explore more faster exact inversion techniques. Similarly, we will try to extend these concepts beyond DDIM-based diffusion models to other generative frameworks.
{
    \small
    \bibliographystyle{ieeenat_fullname}
    \bibliography{main}
}


\end{document}